\documentclass[letterpaper,10pt,conference]{ieeeconf}  %

\IEEEoverridecommandlockouts                              

\usepackage{epsfig} 
\usepackage{times} 
\usepackage{amsmath} 
\usepackage{amssymb}  
\usepackage{amsfonts}
\usepackage{booktabs}
\usepackage{algorithm}
\usepackage{algpseudocode}
\usepackage{bm}  
\usepackage{url}
\usepackage[dvipsnames]{xcolor}
\usepackage{multirow}
\usepackage{comment}
\usepackage{subcaption}
\usepackage{cite}

\usepackage{xurl}
\usepackage{pifont}

\newcommand{\cmark}{\ding{51}} 
\newcommand{\xmark}{\ding{55}} 

\definecolor{commentGreen}{rgb}{0,0.5,0.05}

\title{RoboVAD: A Large Cross-Domain Evaluation Benchmark for Anomaly Detection in Robotic Arm Manipulation Videos\vspace{-0.3cm}}

\author{%
Alexandru-Bogdan Dura$^{1,2}$, Sebastian Balmu\c{s}$^{1,2}$, Radu Tudor Ionescu$^{2}$\\
\hspace{-0.1em}$^{1}$ICI Bucharest, Romania\hspace{0.4em} $^{2}$University of Bucharest, Romania\\
{\tt\small bogdan.dura@ici.ro, sebastian.balmus@ici.ro, raducu.ionescu@gmail.com}\vspace{-0.5cm}
\thanks{This work was supported by a grant of the Ministry of Research, Innovation and Digitization, CNCS -
UEFISCDI, project number PN-IV-P1-PCE-2023-0354, within PNCDI IV.}
}

\begin{document}
\bstctlcite{IEEEexample:BSTcontrol}
\renewcommand{\dbltopfraction}{0.98}
\renewcommand{\textfraction}{0.02}

\maketitle
\thispagestyle{empty}
\pagestyle{empty}

\begin{abstract}
Video anomaly detection (VAD) is an actively studied task, having wide applications in typical scenarios such as public surveillance and road traffic safety. The task is also relevant for robotic arm interactions, where it has several downstream applications, including learning better interaction and manipulation abilities, triggering recovery procedures when anomalies occur, etc. Despite its relevance, the exploration of anomaly detection in robotic arm manipulation videos is limited by the low number of available resources. To this end, we introduce RoboVAD, a large-scale benchmark for video anomaly detection that comprises challenging cross-domain evaluation scenarios, where certain actions (tasks executed by a robotic arm) and anomaly types (mistakes that occur while performing certain tasks) remain unseen during training. RoboVAD is designed to benchmark VAD methods in realistic scenarios, where robotic arms can perform unforeseen tasks, and thereby encounter new anomaly types. We train and evaluate several state-of-the-art VAD methods, including a novel method specifically adapted for robotic arm manipulation. While the proposed method outperforms many state-of-the-art competitors, all methods remain below a micro-averaged frame-level AUC threshold of $70\%$ in the most challenging evaluation setup, confirming the difficulty of the proposed benchmark. We publicly release our dataset and code at \url{https://zenodo.org/records/22754659}.
\end{abstract}


\setlength{\abovedisplayskip}{2.0pt}
\setlength{\belowdisplayskip}{2.0pt}


\section{Introduction}
\vspace{-0.1cm}

In spite of the growing interest in video anomaly detection (VAD) \cite{Madan-TPMAI-2024,nafez2025frameshield,Pang-CSUR-2021,ramachandra2022survey,rolland2026fidel,wu2023vadclip,yang2025monitor}, the task remains challenging because anomalous events are rare, context-dependent, and difficult to enumerate in advance \cite{Huang-TM-2022,ionescu2019objectcentric}. In the context of robotic arm manipulation, a manipulation outcome that is expected in one task may indicate a failure in another. For example, releasing an object is normal when placing it inside a container, but abnormal when the object is dropped before reaching the target. The dependence on context gives rise to a potentially unbounded set of anomalies and makes it impractical to collect representative training examples for every failure that a robotic system may encounter. As a result, most VAD methods learn a model of normality from anomaly-free training videos \cite{Deng-CVPR-2022,Morais-CVPR-2019,Ionescu-WACV-2019,Wu-ECCV-2022}, while alternative weakly supervised approaches use abnormal videos with coarse (video-level) annotations \cite{tian2021rtfm,zaheer2020claws,zhou2023urdmu}.

The aforementioned formulations offer complementary advantages. Normal-only methods preserve the open-set nature of anomaly detection, since any sufficiently large deviation from the learned normal behavior can be detected as abnormal. However, such methods cannot exploit examples of known failures during training. Conversely, supervised or weakly supervised methods can learn more discriminative representations from abnormal examples, but they are commonly evaluated in a closed-set scenario, where the failure categories observed during testing are also represented in the training set. Such an evaluation does not measure whether a model can detect unforeseen failures, even though this capability is essential for robots operating outside tightly controlled environments. Following previous work on VAD in public surveillance videos \cite{acsintoae2022ubnormal,wu-CVPR-2024}, we consider cross-domain evaluation protocols in which manipulation tasks and failure categories in training and test sets are disjoint.

While VAD has been extensively investigated in public surveillance \cite{Georgescu-CVPR-2021,Wang-CVPR-2026,Xu-CVPR-2026,zaheer2022gcl}, road traffic monitoring \cite{Doshi-CVPR-2021,Xing-CVPR-2025,yao2023dota}, and related settings, anomaly detection in robotic arm manipulation videos remains insufficiently explored. Existing resources for robotic failure detection are typically limited to a small number of tasks, narrow collections of failure categories or specialized sensory configurations involving depth, force-torque measurements or proprioception \cite{inceoglu2021finonet,rolland2026fidel,sliwowski2025conditionnet,thoduka2024handover,thoduka2021visual}. Moreover, several existing benchmarks focus on failure classification, semantic diagnosis or execution-level success prediction rather than the frame-level temporal localization of anomalous events. These differences limit the direct applicability of existing benchmarks to low-cost robotic systems that must visually identify when a failure begins and ends.

\begin{figure}[t]
  \centering

  \includegraphics[width=0.92\linewidth]{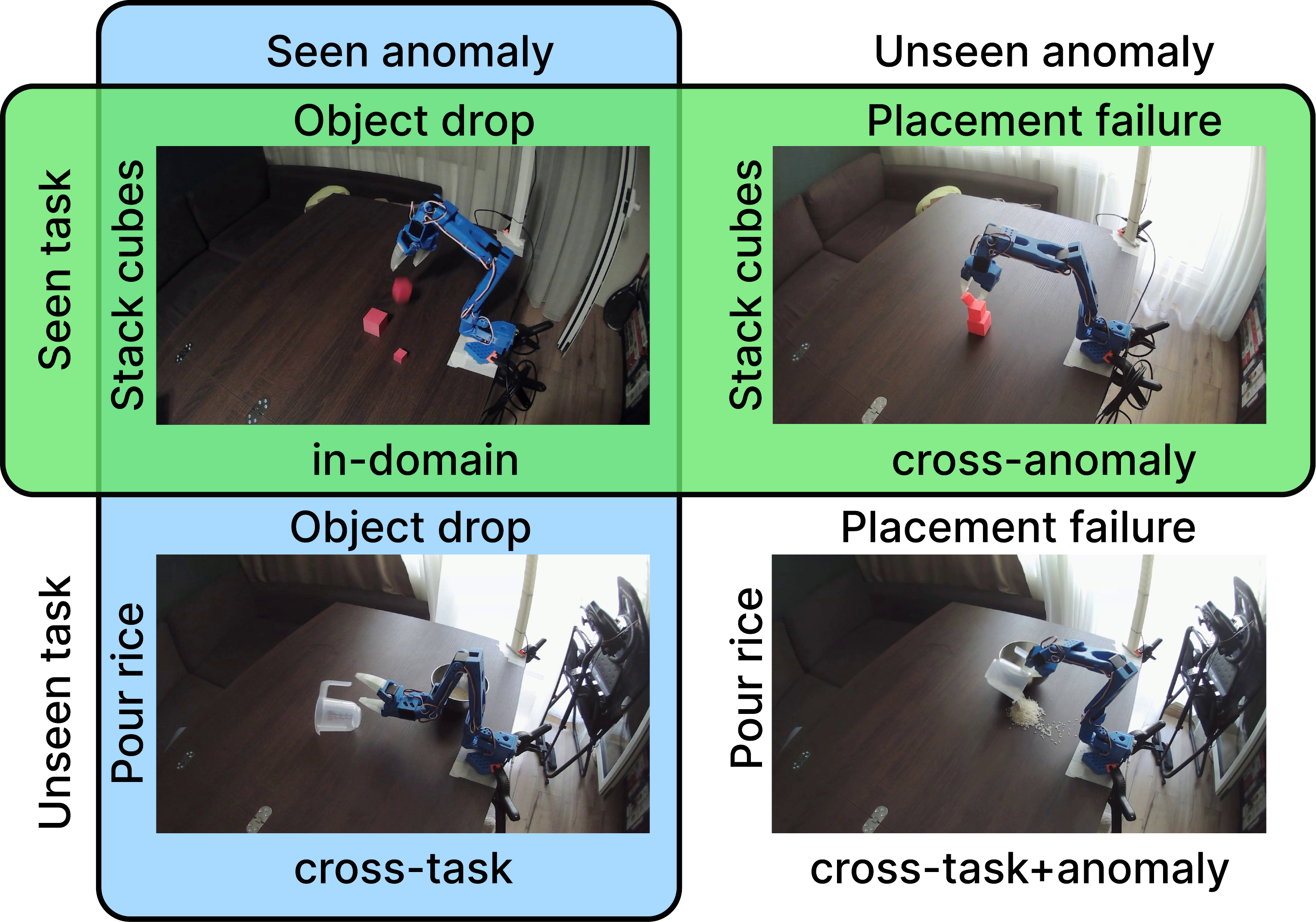} 

  \vspace{-0.1cm}
  \caption{RoboVAD evaluation settings across seen and unseen tasks, as well as seen and unseen anomaly categories, defined relative to the training partition.}
  \label{fig:teaser}
\vspace{-0.45cm}
\end{figure} 

Detecting such events directly from video has important implications for autonomous manipulation. A reliable detector can trigger recovery procedures, prevent a robot from propagating an early error through the remaining execution, and provide informative failure segments for learning improved manipulation policies. However, a model deployed in the real world cannot be expected to operate only on the tasks, objects, camera conditions, and failure types encountered during training. For instance, a household or service robot may be required to perform a new manipulation task, and the corresponding failures may differ substantially from those observed in its training data. Hence, an evaluation restricted to familiar tasks and familiar anomaly types can provide an overly optimistic estimate of practical performance. Yet, none of the existing benchmarks consider realistic evaluation scenarios, where manipulation tasks and anomaly categories can diverge between training and testing, as illustrated in Figure \ref{fig:teaser}.

To address this limitation, we propose \textbf{RoboVAD}, a large-scale benchmark for cross-domain anomaly detection in robotic arm manipulation videos. 
RoboVAD comprises 620,979 frames, corresponding to 5.75 hours of video, including 34,900 anomalous frames and 1,570 temporally localized anomaly events. The dataset covers five manipulation tasks: single-cube pick-and-place, multiple-cube pick-and-place, cube stacking, rice pouring, and ring insertion. The annotated anomalies belong to five categories: failed grasp, collision, placement failure, object drop, and object not dropped.
A central characteristic of RoboVAD is its specific data organization for cross-domain evaluation. We consider the manipulation task and the anomaly category as two distinct sources of domain variation. In the cross-task setting, models are evaluated on rice pouring and ring insertion, after being trained on cube pick-and-place and stacking tasks. In the cross-anomaly setting, models are tested on failure categories that are absent from the training split. Combining these two dimensions produces the most challenging scenario (cross-task+anomaly), in which both the manipulation tasks and the associated anomaly types are unseen. This setup simulates a realistic deployment condition, where a model is trained using the tasks and failures available during training, but is subsequently expected to monitor new robot behaviors and detect mistakes that could not have been anticipated or exhaustively collected beforehand.

We organize the benchmark into in-domain and cross-domain protocols, which makes it possible to separately analyze generalization across manipulation tasks, anomaly types, or both. We evaluate a diverse collection of methods drawn from robotic failure detection, self-supervised VAD, weakly supervised VAD, and language-guided anomaly detection. We additionally introduce two models based on optical-flow representations: (i) FlowJigsaw-SSL, which learns normal motion through multiple self-supervised proxy tasks, and (ii) FlowJigsaw-Supervised, which learns from normal executions and a subset of annotated failures. Although several methods achieve strong results in the in-domain setting, performance decreases considerably when either the task domain or the anomaly categories are changed. In the cross-task+anomaly protocol, all evaluated approaches remain below a frame-level micro-averaged AUC of $70\%$, demonstrating that RoboVAD presents substantial challenges that are not captured by conventional in-domain evaluation. This result highlights the need for more robust and generalizable anomaly detection methods for robotic manipulation.

In summary, our contributions are as follows:
\begin{itemize}
\item We introduce RoboVAD, a large-scale realistic benchmark for anomaly detection in robotic arm manipulation videos, comprising 620,979 frames, 1,570 temporally localized anomalies, two synchronized camera views, and various manipulation tasks and anomaly categories.
\item We propose a comprehensive evaluation protocol that jointly studies cross-task generalization and cross-anomaly detection, enabling models to be tested on unseen manipulation tasks, failure types, and both.
\item We benchmark a representative collection of methods, providing a broad empirical comparison under consistent frame-level evaluation protocols.
\item We introduce FlowJigsaw-SSL and FlowJigsaw-Supervised, two motion-oriented models designed to capture deviations from expected robotic manipulation dynamics under normal-only and supervised training, respectively.
\item We show that current methods experience a substantial performance degradation under cross-domain evaluation, with all evaluated approaches remaining below $70\%$ micro-averaged frame-level AUC when both manipulation tasks and anomaly types are unseen.
\end{itemize}

\section{Related Work}
\label{Sec:related work}

\subsection{Video Anomaly Detection}

Video anomaly detection (VAD) focuses on temporally localizing events that deviate from expected behavior. Due to the fact that anomalous events are rare, most methods learn a representation of normality from anomaly-free training data alone \cite{Pang-CSUR-2021,ramachandra2022survey}, treating any significant deviation at test time as an anomaly. Early approaches used autoencoders to detect deviations through reconstruction errors \cite{hasan2016temporal}, while prediction-based methods compared predicted future frames with their observed counterparts \cite{liu2018future}. Later, object-centric methods suppressed background noise by focusing feature learning strictly on detected objects~\cite{ionescu2019objectcentric,bergaoui2022objectcentric}. Other approaches explicitly modeled the relationship between appearance and motion, either by learning decoupled representations or by measuring their correspondence~\cite{Li-CVIU-2021,Nguyen-ICCV-2019}. Optical flow has also been incorporated through motion reconstruction and flow-guided frame prediction~\cite{georgescu2021background,Liu-ICCV-2021}.

Recent frameworks leverage self-supervised auxiliary tasks to model spatio-temporal structures, including temporal ordering, event completion, and jigsaw puzzle solving~\cite{Georgescu-CVPR-2021,Luo-Arxiv-2020,wang2022jigsaw,zhang2024multiscale}. Particularly, Georgescu et al.~\cite{Georgescu-CVPR-2021} introduced a multi-task framework combining arrow-of-time prediction, motion irregularity detection, appearance reconstruction, and knowledge distillation. Similarly, Wang et al.~\cite{wang2022jigsaw} learned nominal representations by solving decoupled spatial and temporal jigsaw permutation tasks, while Barbalau et al.~\cite{barbalau2023ssmtlpp} scaled the multi-task paradigm using modern backbones, refined region filtering, and extra proxy objectives, including jigsaw puzzle solving and inpainting. Masked reconstruction has also been combined with motion-weighted learning and self-distillation to improve the detection of localized anomalies~\cite{ristea2024masked}.

Beyond normal-only learning, weakly supervised methods use video-level labels to localize anomalous snippets \cite{zaheer2020claws, feng2021mist}. RTFM learns discriminative temporal feature magnitudes~\cite{tian2021rtfm}, UR-DMU models normal and abnormal prototypes~\cite{zhou2023urdmu}, while recent methods such as VadCLIP and DSANet exploit vision-language alignment~\cite{wu2023vadclip,yin2026learning}. Foundation models have additionally enabled training-free or zero-shot detection through caption-based reasoning, online scoring, and user-defined anomaly descriptions~\cite{zanella2024lavad,yang2025monitor,ahn2026anyanomaly}.

\begin{table*}[t]
\caption{Comparison of robotic arm anomaly detection datasets. Existing datasets typically contain a small number of anomaly instances (mostly under 150), while temporal (frame-level) annotations are missing for larger datasets. Compared with existing datasets, RoboVAD is the only cross-domain evaluation benchmark and contains $3\times$ more anomaly instances than all other datasets.}
\label{tab:datasetcomparison}
\vspace{-0.3cm}
\setlength\tabcolsep{0.33em}
\begin{center}
  \begin{tabular}{|l|c|cccccc|c|c|c|c|c|c|c|}
  \hline
    \multirow{2}{*}{Dataset} & Length & \multicolumn{6}{c|}{Number of frames} & Annot. & Anomaly & Anomaly & Robotic & Proprio- & Cross & Cross \\
    \cline{3-8}
     & (h) & Total & Train & Val & Test & Normal & Abnormal & level & instances & categories & tasks & ception & task& anomaly \\
    \hline
    \hline
    ImperfectPour \cite{sliwowski2025conditionnet}  & 1.22 & 131,476 & 102,731 & - & 28,745 & 45,240 & - & video & 86 & 3 & 4  & \xmark & \xmark & \xmark \\ 
    FAILURE \cite{inceoglu2021finonet} & 5.55 & 17,400 & N/A & - & N/A & N/A & - & video & 147 & 1 & 5  & \xmark & \xmark & \xmark \\
    Handover \cite{thoduka2024handover}  & 2.37 & 218,503 & 141,462 & 30,935 & 46,106 & 54,569 & - & video & 433 & 4 & 2  & \cmark & \xmark & \xmark \\ 
    \hline
    Botfails \cite{rolland2026fidel}  & 3.84 & 414,359 & \textbf{232,289} & - & 182,070 & 352,954 & \textbf{60,210} & frame & 129 & 1 & \textbf{10}  & \cmark & \xmark & \xmark \\ 

    Bookshelf \cite{thoduka2021visual} & 1.38 & 49,814 & 20,252 & 2,258 & 27,304 & 46,483 & 3,331 & frame & 66 & \textbf{7} & 1  & \cmark & \xmark & \xmark \\
    
    \hline
    RoboVAD (Ours)  & \textbf{5.75} & \textbf{620,979} & 163,445 & \textbf{34,266} & \textbf{423,268} & \textbf{586,079} & 34,900 & frame & \textbf{1,570} & 5 & 5  & \cmark & \cmark & \cmark \\
    
    \hline
    
    \end{tabular}
\end{center}
\label{table:datasetcomparison}
\vspace{-0.5cm}
\end{table*}

Open-set evaluation considers whether detectors generalize beyond the anomaly categories observed during training. UBnormal formalized a supervised protocol with disjoint training and evaluation anomaly classes~\cite{acsintoae2022ubnormal}, while LaGoVAD introduced language-guided open-world detection with adaptable anomaly definitions~\cite{liu2026languageguided}. Following these formulations, our benchmark investigates both normal-only self-supervised learning and supervised open-set detection. Unlike existing open-set VAD benchmarks, we focus on detecting anomalies in video recordings of robotic arm manipulations, which represents a fundamentally different setup than typical video surveillance datasets. More specifically, RoboVAD contains subtle domain-specific anomalies, which may be regarded as irrelevant in the context of typical setups, e.g.~public video surveillance \cite{lu2013abnormal,Ramachandra-WACV-2020a}, road traffic \cite{yao2023dota}, etc.

\subsection{Robotic Manipulation Failure Detection and Datasets}

Failures in robotic manipulation include unsuccessful grasps, dropped objects, collisions, incorrect placements, and incomplete executions. Although failures can be detected as visual anomalies, the two concepts are not equivalent: an unusual observation may be benign, while some task failures may be visually subtle. FIDeL~\cite{rolland2026fidel} is a method  that explicitly addresses this distinction by combining anomaly detection with semantic filtering to separate benign deviations from failures that threaten task completion. In our benchmark, the annotations represent physical manipulation failures, while visual anomaly detection provides the learning formulation used to localize them.

Early failure-detection methods combined visual, auditory, haptic, and kinematic observations to model nominal robot executions~\cite{park2016multimodal}. FINO-Net later fused RGB, depth and audio to detect and classify failures in tabletop manipulation~\cite{inceoglu2021finonet,inceoglu2024multimodal}. More closely related to our setting, Thoduka et al.~\cite{thoduka2021visual} learned nominal optical-flow dynamics for book placement and accounted for camera and manipulator motion using robot kinematics. ConditionNET instead learns action-conditioned preconditions and effects, detecting failures when observations do not satisfy the expected outcome of an action~\cite{sliwowski2025conditionnet}.

Existing datasets differ considerably in their modalities and annotation granularity. The Bookshelf dataset provides frame-level anomaly labels for a single placement task~\cite{thoduka2021visual}, while FAILURE supports multimodal detection and classification across several tabletop actions~\cite{inceoglu2021finonet,inceoglu2024multimodal}. The Handover Failure Detection dataset targets failures during human-robot object exchange and combines video with robot-state and force-torque information~\cite{thoduka2024handover}. (Im)PerfectPour provides multi-view demonstrations with execution outcomes and temporal action phases~\cite{sliwowski2025conditionnet}, whereas REASSEMBLE focuses on multimodal, contact-rich assembly and disassembly with hierarchical action and success annotations~\cite{sliwowski2025reassemble}.

Larger recent benchmarks address complementary objectives. ARMBench studies object-centric perception and robot-induced defects in warehouse manipulation~\cite{mitash2023armbench}. BotFails combines multi-view video, proprioception, and language across varied tasks, while distinguishing benign anomalies from genuine failures~\cite{rolland2026fidel}. ViFailback focuses on failure keyframes, explanations, and corrective guidance through vision-language supervision~\cite{zeng2026vifailback}. In contrast, our benchmark is designed for frame-level localization of physical failures in low-cost, multi-view tabletop manipulation. It provides temporally localized failure annotations and supports normal-only, supervised open-set, cross-task, and camera-view evaluation protocols. As shown in Table \ref{table:datasetcomparison}, RoboVAD is the only dataset in the field that considers challenging cross-domain evaluation setups, including cross-task and cross-anomaly, while also containing $3\times$ more anomaly instances than existing benchmarks.

\section{RoboVAD Benchmark}
\label{Sec:RoboVAD Benchmark}

\noindent
\textbf{Overview and motivation.} RoboVAD is a benchmark for frame-level video anomaly detection in real-world robotic manipulation. Detecting 
manipulation errors can support recovery procedures and provide informative examples for learning more reliable interactions. The dataset comprises 1,078 execution episodes recorded from two camera viewpoints, covering five manipulation tasks and five anomaly categories. Temporal annotations identify individual anomalous events within each episode, including multiple events that may occur during the same execution. Beyond evaluation on familiar tasks and failure types, RoboVAD provides cross-task, cross-anomaly and combined cross-task+anomaly evaluation protocols. These settings assess whether detectors can recognize failures when the manipulation task, anomaly category or both are unseen during training.

\noindent
\textbf{Setup and tasks.} We collect real-world recordings of robotic manipulation in a tabletop environment using an SO-ARM101 robot and the LeRobot framework \cite{cadenelerobot}. Data collection uses a leader–follower teleoperation setup: a human operator physically moves the leader arm, while the follower arm reproduces these movements to manipulate objects in the workspace. Two static cameras are mounted at the edges of the table at slightly different heights and viewing angles, providing complementary views of the same execution. Both cameras record at 30 FPS with a resolution of \(1920 \times 1080\) pixels. The robot's structural components, camera mounts, and most task props are 3D-printed in PLA. For each episode, RoboVAD provides recordings from both camera views together with the robot state and action data recorded through LeRobot, including joint and gripper positions. It also provides frame-level anomaly annotations and episode assignments to the training, validation and test sets. 

\noindent
\textbf{Tasks.} The dataset covers five manipulation tasks: \emph{single-cube pick-and-place}, \emph{multiple-cube pick-and-place}, \emph{cube stacking}, \emph{rice pouring}, and \emph{ring insertion}. The \emph{pick-and-place tasks} require transferring one or multiple cubes to a target container. In the \emph{stacking task}, the robot builds a tower from three cubes of different sizes, with the largest cube at the bottom and the smallest at the top. \emph{Rice pouring} requires transferring rice from a cup into a bowl, while \emph{ring insertion} requires placing three rings onto a peg. Together, these tasks involve object grasping and transport, ordered placement, alignment, and manipulation of granular material. 

\begin{figure*}[t]
  \centering
  \setlength{\tabcolsep}{3pt}
  
  \includegraphics[width=\textwidth]{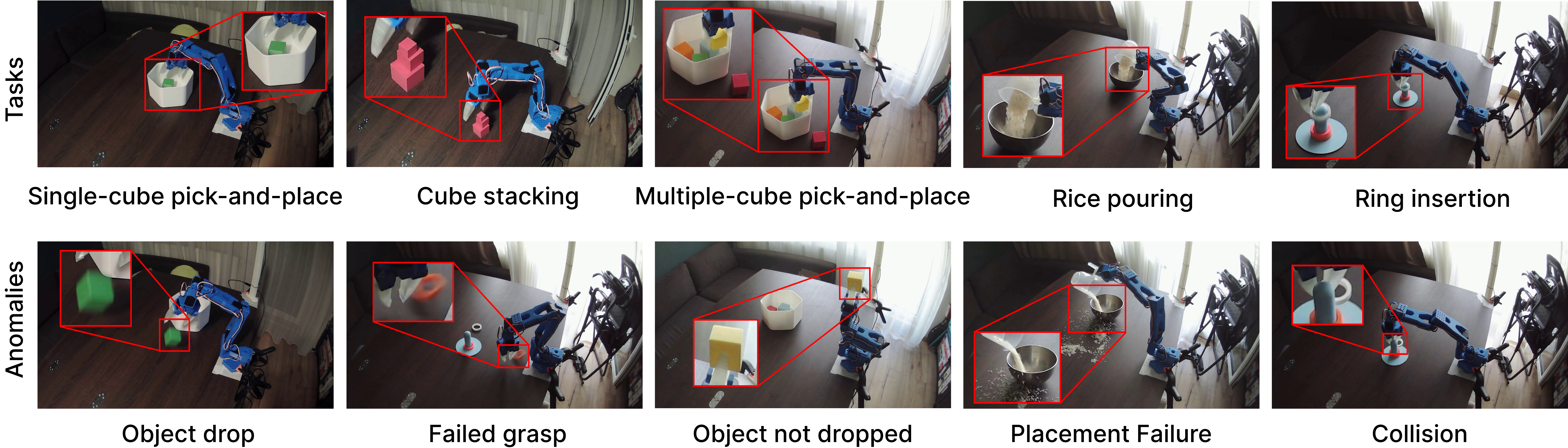} 

  \caption{Example frames illustrating the five manipulation tasks (top) and five anomaly categories (bottom) in RoboVAD.}
  \label{fig:robovad_tasks_anomalies}
\vspace{-0.45cm}
\end{figure*} 

\noindent
\textbf{Anomaly types.}
The variety of tasks inherently induces different types of potential anomalies. We consider normal manipulation to be an execution in which the robot grasps, transports and releases objects as intended, progressing toward the task goal. An anomaly is an observable deviation from the intended interaction between the robot, the manipulated objects and the environment. We distinguish five categories of anomalies: \textit{failed grasp}, when the robot attempts to grasp an object but fails to establish a stable grip; \textit{collision}, when the robot or the object that is carried makes unintended contact with another object or the environment; \textit{object drop}, when an object that is carried slips from the gripper and falls; \textit{placement failure}, when an object is released in the target area but does not reach or remain in its intended position; \textit{object not dropped}, when an object remains in the gripper despite an intended release. An episode may contain multiple anomalous events and still achieve the task goal. In Figure \ref{fig:robovad_tasks_anomalies}, we present video frames illustrating the manipulation tasks and various types of anomalies included in RoboVAD. 

\noindent
\textbf{Episodes.} Each episode corresponds to one task execution. Recording begins when the robot starts moving, and ends when the task is completed or no further action can advance its completion. Episodes are not terminated at the first anomaly, execution continues, while further progress remains possible. Consequently, a single episode may contain multiple anomalous events. 

\noindent
\textbf{Preprocessing and annotation.} Episodes are temporally trimmed to retain the task execution, from the first robot movement to the final action. A short margin of approximately 5 to 10 frames may remain before and after the execution. The anomalies are annotated at the frame level. Anomalous events are assigned to one of five categories: \emph{failed grasp}, \emph{collision}, \emph{placement failure}, \emph{object drop} or \emph{object not dropped}. Each event is additionally annotated with its severity, classified as major or minor, and its visibility across the two camera views. We consider major anomalies to be events that require the robot to perform additional actions to complete the intended manipulation, such as an object drop that requires the robot to retrieve and grasp the object again before continuing the task. Annotated anomaly intervals do not overlap, although consecutive anomalies may occur without a normal interval in between. The dataset includes anomalies resulting from teleoperation errors as well as deliberately induced events. 

All recordings were collected and annotated by the authors under standardized criteria for anomaly taxonomy, temporal boundaries, visibility and severity. To validate reliability, two authors independently annotated a 50 episode subset, achieving Cohen's $\kappa=1.00$ for episode-level detection, $\kappa=0.79$ for frame-level anomaly labels, and $\kappa=0.71$ for anomaly categories. All labels were cross-reviewed and disagreements were resolved through discussion. 

\noindent
\textbf{Dataset splits.} We split the dataset into training, validation and test sets at the episode level, keeping both camera views and all annotations from an episode in the same partition. Single-cube pick-and-place, multiple-cube pick-and-place and cube stacking constitute the source tasks, while rice pouring and ring insertion are reserved for testing generalization to unseen tasks. Within the source tasks, anomalous episodes are assigned according to their annotated failure categories. Episodes containing placement failure or collision are assigned to the test set. Among the remaining episodes, those containing objects not dropped are assigned to validation, while those containing object drop or failed grasp are assigned to training. A subset of episodes containing failure categories seen during training is reserved for testing. All episodes from the two unseen tasks are also assigned to the test set. This construction supports evaluation on both seen and unseen tasks, with failure categories that are either represented in or absent from the training set. The corresponding evaluation settings are detailed in the experiments section. The dataset comprises 1,078 episodes, including 616 normal and 462 anomalous episodes, split into 367 training, 82 validation and 629 test episodes. The in-domain test set includes 259 episodes, the cross-task set 308 episodes, the cross-anomaly set 297 episodes and the cross-task+anomaly set 283 episodes. Within each task group, normal episodes and episodes containing both seen and unseen anomaly categories contribute to both settings. In these mixed episodes, normal frames are retained in both settings, while anomaly frames are selected according to the evaluated categories. 




\noindent
\textbf{Evaluation protocol.} We evaluate all methods under four settings that distinguish generalization across manipulation tasks and anomaly categories. The in-domain setting evaluates failure categories represented in training on the source tasks: \textit{single-cube pick-and-place}, \textit{multiple-cube pick-and-place} and \textit{cube stacking}. The cross-anomaly setting evaluates unseen failure categories on the same tasks. The cross-task setting evaluates anomaly categories represented in training on the two held-out tasks, \textit{rice pouring} and \textit{ring insertion}. Finally, the cross-task+anomaly setting evaluates unseen failure categories on the held-out tasks. Throughout these settings, ``seen'' and ``unseen'' refer to the contents of the training partition, irrespective of the supervision used by an individual method. For each setting, we include normal frames from the corresponding tasks and anomalous frames belonging to the selected failure categories. Frames annotated with other failure categories are excluded from that evaluation. 

\noindent
\textbf{Evaluation measures.} We choose the area under the ROC curve (AUC), computed using the frame-level anomaly scores and ground-truth annotations. Following \cite{georgescu2021background}, we report both micro and macro AUC scores. Micro AUC is computed by pooling the evaluated frames from all episodes, while macro AUC is computed separately for each episode and then averaged, giving equal weight to each episode. 

\section{Methods}
\label{sec:methods}

We investigate frame-level anomaly detection as a means of identifying and temporally localizing execution failures in robotic manipulation videos. Given a video of a manipulation episode, the task is to assign each frame an anomaly score that reflects the likelihood of an execution failure occurring at that point in time. We consider approaches from robotic failure detection, and video anomaly detection, as well as a multi-task self-supervised model and a supervised counterpart built on the same backbone.

\noindent\textbf{Robotic failure detection.}
As a domain-specific baseline, we employ FIDeL~\cite{rolland2026fidel}, a framework designed to identify failures during robotic policy execution. FIDeL first measures visual deviations from expert demonstrations using optimal transport alignment. It then applies conformal thresholding to determine whether these deviations indicate a failure and uses a vision-language model to filter detections according to their semantic relevance. This baseline assesses the performance of an approach developed specifically for robotic behavior monitoring.


\noindent\textbf{Weakly supervised anomaly detection.}
The weakly supervised baselines include UR-DMU~\cite{zhou2023urdmu}, VadCLIP~\cite{wu2023vadclip}, and DSANet~\cite{yin2026learning}. These methods learn to localize anomalous events using video-level labels and therefore do not require temporal failure annotations. UR-DMU represents normal and abnormal patterns using dual memory units and regulates their separation through uncertainty learning. VadCLIP adapts CLIP using a visual classification branch and a vision-language alignment branch. DSANet combines normality modeling with disentangled semantic alignment. Together, these approaches provide a comparison with models that learn from abnormal training videos under weak supervision.

\noindent\textbf{Language-guided anomaly detection.}
For language-guided anomaly detection, we employ LaGoVAD~\cite{liu2026languageguided}. The framework conditions its predictions on natural-language definitions of anomalous events supplied at inference time. As a result, the anomaly criterion can be adapted to the task without retraining the model. This property is relevant to robotic manipulation, where the same observed event may be either expected or erroneous depending on the intended action.

\noindent\textbf{Self-supervised anomaly detection.}
As a self-supervised baseline, we employ Jigsaw-VAD~\cite{wang2022jigsaw}. The framework learns representations of normal video patterns by solving spatial and temporal jigsaw puzzles and detects anomalies based on deviations from these learned patterns. Its architecture also serves as the basis for our multi-task model, enabling a direct comparison between different proxy-task formulations for learning normal behavior.

\noindent\textbf{Multi-task self-supervised model.}
Our multi-task model, FlowJigsaw-SSL, builds on the architecture of Jigsaw-VAD~\cite{wang2022jigsaw} and incorporates the proxy tasks introduced in SSMTL++~\cite{barbalau2023ssmtlpp}. The model is motivated by the observation that manipulation failures, including object slips, drops, and collisions, often manifest as deviations from expected motion. Optical-flow clips are therefore used to learn spatio-temporal representations through three complementary tasks: arrow-of-time prediction, motion irregularity detection, and inpainting.

Dense optical flow is computed using the method of Farneb\"ack et al.~\cite{farneback}. Each flow field comprises the horizontal displacement $u$, the vertical displacement $v$, and the motion magnitude $m=\sqrt{u^2+v^2}$. For episodes recorded from two camera views, the corresponding flow fields are concatenated along the channel dimension. The resulting clip is denoted by $X\in\mathbb{R}^{T\times H\times W\times 6}$, where $T$ is the number of flow fields and $H$ and $W$ are their spatial dimensions. A shared 3D convolutional backbone extracts features from each clip. The backbone consists of convolutional layers, instance normalization, ReLU activations, and spatial max-pooling, followed by three task-specific heads.

For arrow-of-time prediction~\cite{Wei-CVPR-2018}, clips are presented in either their original or reversed temporal order, and the first head predicts whether the temporal order is correct. For motion irregularity detection, the input retains its original temporal structure or is artificially disturbed, and the second head predicts whether the resulting motion is regular. Both heads are trained using binary cross-entropy with logits. Their respective losses are denoted by $\mathcal{L}_{\mathrm{AoT}}$ and $\mathcal{L}_{\mathrm{MI}}$.

For the inpainting task, which follows the context-encoding principle introduced by Pathak et al.~\cite{Pathak-CVPR-2016}, each frame is divided into a $4\times4$ grid, and patches covering 25\% of the frame area are masked. The third head reconstructs the missing regions from the remaining spatial and temporal context. Let $\widehat{X}_b$ denote the reconstruction of sample $b$, and let $M_{b,j}$ equal one for masked entries and zero otherwise. For a batch of $B$ clips, the reconstruction loss is defined as
\begin{equation}
\mathcal{L}_{\mathrm{IP}}
=
\frac{1}{B}\sum_{b=1}^{B}
\frac{\sum_j M_{b,j}\left(X_{b,j}-\widehat{X}_{b,j}\right)^2}
{\sum_j M_{b,j}},
\label{eq:inpainting}
\end{equation}
where $j$ indexes the temporal, spatial, and channel dimensions. The shared backbone and the three task-specific heads are trained jointly using:
\begin{equation}
\mathcal{L}_{\mathrm{total}}
=
\mathcal{L}_{\mathrm{AoT}}
+
\mathcal{L}_{\mathrm{MI}}
+
0.1\,\mathcal{L}_{\mathrm{IP}}.
\label{eq:joint_loss}
\end{equation}

At inference time, the model is applied to clips in their original temporal order. Let $p_{\mathrm{AoT}}(X)$ and $p_{\mathrm{MI}}(X)$ denote the predicted probabilities that clip $X$ has the correct temporal order and contains regular motion, respectively. Let $e_{\mathrm{IP}}(X)$ denote its masked reconstruction error, computed as in Eq.~\eqref{eq:inpainting}, before averaging across the batch. The task-specific anomaly scores are defined as:
\begin{equation}
\begin{aligned}
s_{\mathrm{AoT}}(X) &= 1-p_{\mathrm{AoT}}(X),\\
s_{\mathrm{MI}}(X) &= 1-p_{\mathrm{MI}}(X),\\
s_{\mathrm{IP}}(X) &= e_{\mathrm{IP}}(X).
\end{aligned}
\label{eq:task_scores}
\end{equation}
The three scores are normalized to account for differences in scale and then averaged to obtain the clip-level anomaly score:
\begin{equation}
s(X)
=
\frac{1}{3}\left(
\widetilde{s}_{\mathrm{AoT}}(X)
+
\widetilde{s}_{\mathrm{MI}}(X)
+
\widetilde{s}_{\mathrm{IP}}(X)
\right),
\label{eq:combined_score}
\end{equation}
where the tilde denotes score normalization. The model is evaluated over overlapping clips, and each clip-level score is assigned to all frames within that clip. Scores from clips that overlap at a given frame are averaged to obtain the final frame-level prediction.

\noindent\textbf{Supervised model.}
To measure the benefit of training with abnormal examples, we evaluate FlowJigsaw-Supervised, a supervised variant of the same 3D convolutional backbone. The three proxy-task heads are replaced with a single binary classification head that predicts whether an input clip is normal or anomalous. Following the open-set protocol of UBnormal~\cite{acsintoae2022ubnormal}, the model is trained on normal executions, as well as the subset of anomaly categories provided in the training set.
Since failure categories used for training and evaluation are disjoint, this setting assesses whether supervision from known failures generalizes to unseen failure types.

\section{Experiments}
\label{Sec:experiments}

\subsection{Implementation Details}


We adapt the available official implementations of the chosen methods to RoboVAD, retaining their model architectures and introducing dataset-specific loading and evaluation protocols. Frames are initially resized to $128\times128$ pixels and subsequently processed according to each method's requirements. Methods trained exclusively on normal data use the 307 normal training episodes, while supervised and weakly supervised methods use all 367 training episodes along with their respective annotations. 

For DSANet~\cite{yin2026learning}, VadCLIP~\cite{wu2023vadclip} and LaGOVAD~\cite{liu2026languageguided}, we extract frozen CLIP ViT-B/16 features every eight frames, and train the remaining components for 30 epochs with a batch size of 16. The learning rate is $10^{-5}$ for DSANet and VadCLIP and $5\cdot10^{-5}$ for LaGOVAD. For UR-DMU~\cite{zhou2023urdmu}, we extract frozen RGB I3D features and train the remaining components using Adam with a learning rate of $10^{-4}$, weight decay of $5\cdot10^{-5}$ and a batch size of 4. Jigsaw-VAD~\cite{wang2022jigsaw} is trained on normal episodes for 30 epochs using Adam with a learning rate of $10^{-4}$ and a batch size of 128. FrameShield~\cite{nafez2025frameshield} is trained for 15 epochs using AdamW with a learning rate of $10^{-5}$ and a batch size of 8. FIDeL~\cite{rolland2026fidel} constructs normal reference memories from the normal training episodes using a pretrained DINOv2 encoder, whose weights remain fixed.

Both FlowJigsaw variants are trained from random initialization for 15 epochs using Adam with a learning rate of $10^{-4}$ and dropout of 0.3. Inputs contain six optical-flow frames of size $128\times128$, with the horizontal displacement, vertical displacement and motion magnitude from each camera concatenated into six channels. FlowJigsaw-Supervised uses a temporal sampling stride of four frames, a batch size of 64 and weight decay of $10^{-4}$. It is trained with class-weighted focal loss, using $\alpha=0.75$ and $\gamma=2$. Each clip is labeled according to its final frame. FlowJigsaw-SSL uses a temporal sampling stride of 8 frames, a batch size of 32 and no weight decay. At inference, we sum the scores from the three proxy tasks, averaging the reconstruction error over eight random masks. 

\subsection{Anomaly Detection Results}

\begin{table}[t]
\caption{Micro and macro frame-level AUC scores (\%) on RoboVAD across four evaluation settings. Higher values indicate better performance. Best and second-best results on each column are shown in \textcolor{RoyalBlue}{\textbf{bold blue}} and \textcolor{ForestGreen}{green}, respectively.}
\label{tab:results}
\vspace{-0.3cm}
\setlength\tabcolsep{0.09em}
\begin{center}
  \begin{tabular}{|l|c|c|c|c|c|c|c|c|}
  \hline
    \multirow{4}{*}{Method} & \multicolumn{8}{c|}{AUC} \\
     \cline{2-9}
     & \multicolumn{2}{c|}{\multirow{2}{*}{In-domain}} & \multicolumn{2}{c|}{Cross-} & \multicolumn{2}{c|}{\multirow{2}{*}{Cross-task}} & \multicolumn{2}{c|}{Cross-task} \\
     & \multicolumn{2}{c|}{} & \multicolumn{2}{c|}{anomaly} & \multicolumn{2}{c|}{} & \multicolumn{2}{c|}{+anomaly} \\
    \cline{2-9}
    & micro & macro & micro & macro & micro & macro & micro & macro \\
    
    \hline
    \hline
    DSANet \cite{yin2026learning} & 47.3 & 40.7 & 52.1 & 51.0 & 54.4 & 49.8 & 59.7 & 47.8 \\
    FIDeL \cite{rolland2026fidel} & 56.3 & 54.0 & 52.1 & 53.4 & 58.1 & 53.7 & 63.2 & 51.2 \\
    FrameShield \cite{nafez2025frameshield} & 52.6 & 54.8 & 47.8 & 44.0 & 54.3 & 48.6 & 55.1 & 42.0 \\
    Jigsaw-VAD \cite{wang2022jigsaw} & 64.1 & 65.2 & 51.7 & 51.1 & 66.9 & 70.9 & 57.6 & 57.9 \\
    LaGOVAD \cite{liu2026languageguided} & 45.8 & 44.3 & 45.2 & 46.6 & 47.9 & 46.1 & 50.6 & 45.4 \\
    VadCLIP \cite{wu2023vadclip} & 47.1 & 42.0 & 51.5 & 54.1 & 48.0 & 55.1 & 48.0 & 51.9 \\
    UR-DMU \cite{zhou2023urdmu} & \textcolor{ForestGreen}{88.5} & \textcolor{RoyalBlue}{\textbf{94.4}} & 61.3 & 71.0 & \textcolor{ForestGreen}{71.8} & \textcolor{RoyalBlue}{\textbf{77.5}} & 63.9 & \textcolor{RoyalBlue}{\textbf{71.9}} \\
    \hline
    FlowJigsaw-SSL (ours) & 73.6 & \textcolor{ForestGreen}{72.8} & \textcolor{RoyalBlue}{\textbf{79.6}} & \textcolor{RoyalBlue}{\textbf{82.5}} & 65.6 & 64.9 & \textcolor{RoyalBlue}{\textbf{67.4}} & 64.5 \\
    FlowJigsaw-Sup. (ours) & \textcolor{RoyalBlue}{\textbf{94.6}} & \textcolor{RoyalBlue}{\textbf{94.4}} & \textcolor{ForestGreen}{69.0} & \textcolor{ForestGreen}{71.9} & \textcolor{RoyalBlue}{\textbf{72.0}} & \textcolor{ForestGreen}{73.8} & \textcolor{ForestGreen}{66.7} & \textcolor{ForestGreen}{66.8} \\
    
    \hline
    
    \end{tabular}
\end{center}
\label{table:results}
\vspace{-0.1cm}
\end{table}

\begin{table}[t]
\caption{Comparison of RGB and optical-flow (OF) inputs in terms of micro-averaged frame-level AUC (\%). Best results within each model are shown in bold.}
\label{tab:rgbablation}
\vspace{-0.3cm}
\setlength\tabcolsep{0.34em}
\begin{center}
  \begin{tabular}{|l|c|c|c|c|c|}
  \hline
    \multirow{2}{*}{Method} & \multirow{2}{*}{Input} & In- & Cross- & Cross- & Cross-task \\
    & & domain & anomaly & task & +anomaly \\
    \hline
    \hline
    FlowJigsaw-Sup. & RGB & 49.9 & 49.0 & 39.4 & 35.1 \\
    FlowJigsaw-Sup. & OF & \textbf{94.6} & \textbf{69.0} & \textbf{72.0} & \textbf{66.7} \\
    \hline
    FlowJigsaw-SSL & RGB & 50.0 & 43.1 & 46.6 & 32.5 \\
    FlowJigsaw-SSL & OF & \textbf{73.6} & \textbf{79.6} & \textbf{65.6} & \textbf{67.4} \\
    
    \hline
    
    \end{tabular}
\end{center}
\label{table:rgbablation}
\vspace{-0.5cm}
\end{table}

\noindent
\textbf{Quantitative results.} In Table~\ref{table:results}, we report micro- and macro-averaged frame-level AUC across the four evaluation settings. FlowJigsaw-Supervised achieves the highest in-domain micro AUC (94.6\%), exceeding UR-DMU by 6.1\%, while both methods attain a macro AUC of 94.4\%. However, the performance of FlowJigsaw-Supervised is lower when evaluated on unseen anomaly categories. In the cross-anomaly setting, FlowJigsaw-SSL achieves the highest micro and macro AUC scores of 79.6\% and 82.5\%, respectively, outperforming its supervised counterpart by 10.6\% on both micro and macro AUC scores. These results suggest that learning normal motion patterns can support generalization to failure categories absent from training. On unseen manipulation tasks, FlowJigsaw-Supervised obtains the highest micro AUC (72\%), while UR-DMU yields the highest macro AUC (77.5\%). When both tasks and anomalies are unseen, FlowJigsaw-SSL obtains the highest micro AUC (67.4\%), while UR-DMU leads in macro AUC with 71.9\%. All evaluated methods remain below 70\% in terms of micro AUC in the cross-task+anomaly setting, highlighting the difficulty of detecting unfamiliar failures, during new manipulation tasks. 

\begin{figure}[t]
  \centering
  
  \includegraphics[width=0.92\linewidth]{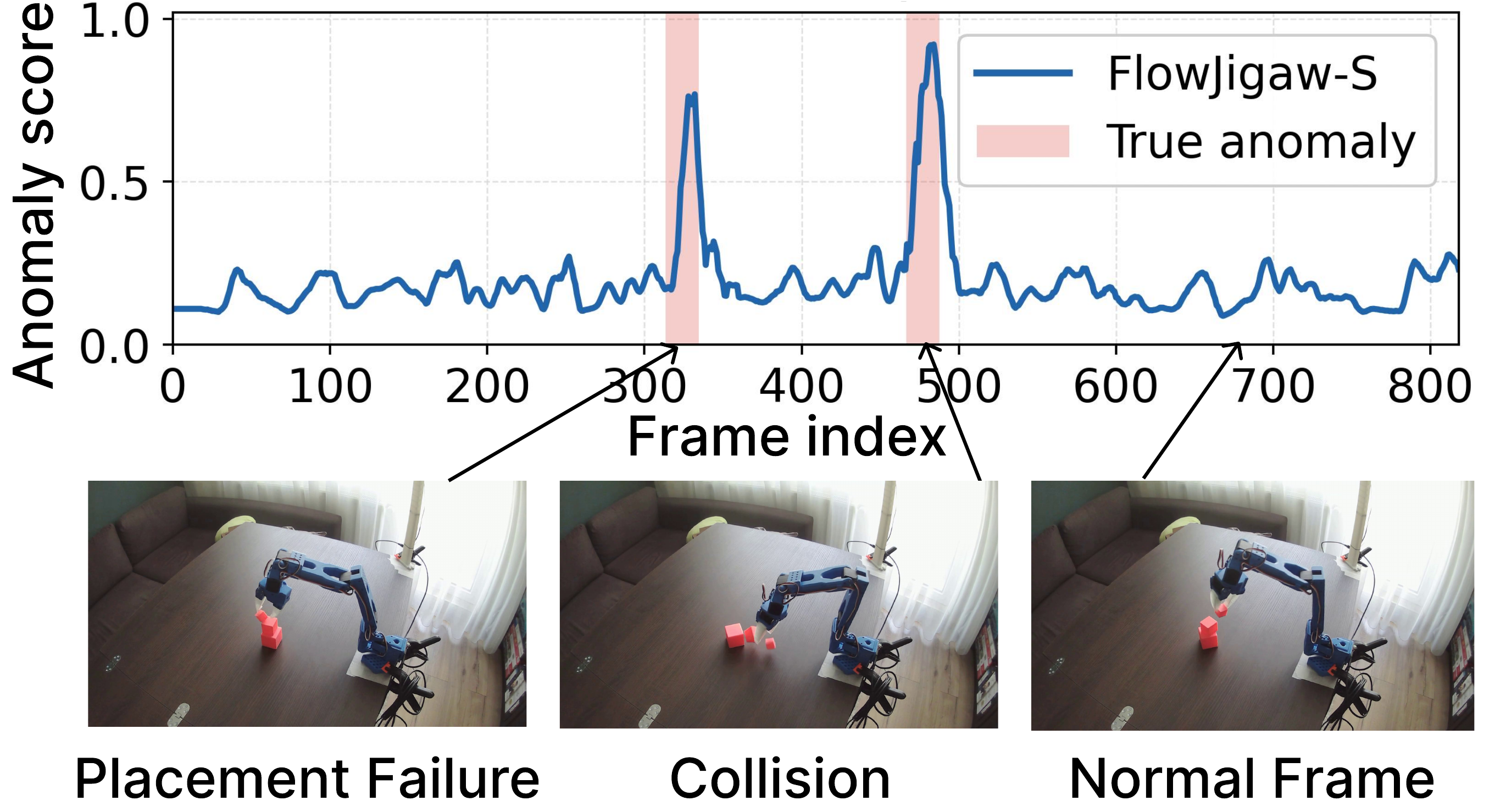} 

  \vspace{-0.15cm}
  \caption{Anomaly scores of FlowJigsaw-Supervised for cube-stacking task episode 119, illustrating cross-anomaly detection of placement failure and collision.}
  \label{fig:qualitative_stackcubes}
\end{figure}

\begin{figure}[t]
  \centering  
  \includegraphics[width=0.92\linewidth]{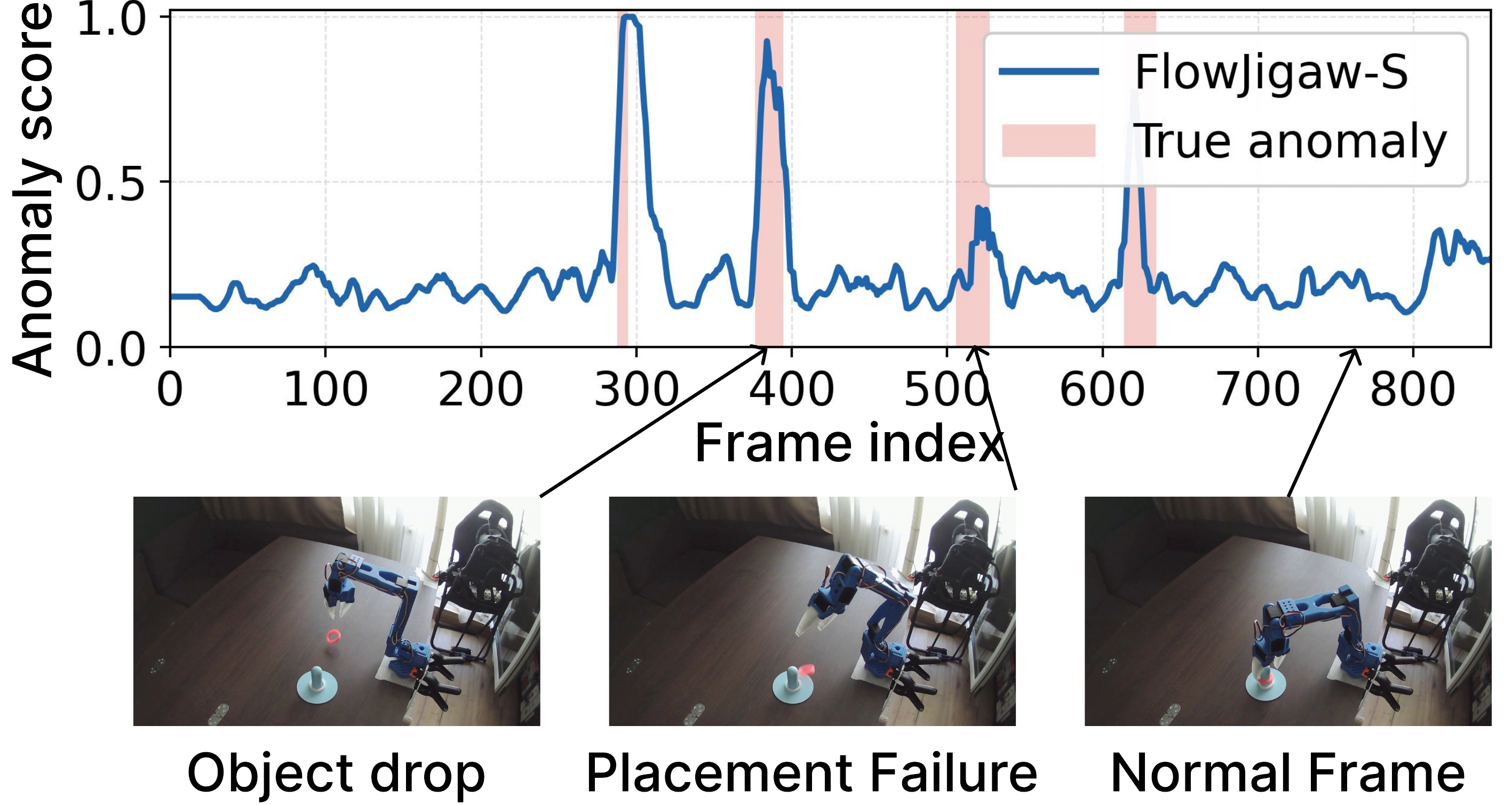} 

  \vspace{-0.15cm}
   \caption{Anomaly scores of FlowJigsaw-Supervised for ring-insertion episode 53, illustrating cross-task detection of object drop and cross-task+anomaly detection of placement failure.}
  \label{fig:qualitative_rings_on_peg}
\vspace{-0.45cm}
\end{figure}

\noindent
\textbf{Qualitative results.} In Figures~\ref{fig:qualitative_stackcubes} and~\ref{fig:qualitative_rings_on_peg}, we show predictions of FlowJigsaw-Supervised for cube stacking (seen task) and ring insertion (unseen task), respectively. In the stacking cubes episode, the pronounced score peaks align with the annotated anomalies. For ring insertion, three annotated anomalous events produce clear score peaks, while the placement failure produces a lower anomaly score. This placement failure occurs in an unseen task and belongs to an unseen anomaly category, illustrating a challenging case of cross-task+anomaly generalization.

\noindent
\textbf{Input representation ablation.} We compare RGB frames and optical flow as inputs to both FlowJigsaw variants. The results shown in Table~\ref{tab:rgbablation} indicate that optical flow consistently improves performance across all four evaluation settings. These results support the use of explicit motion representation for anomaly detection in RoboVAD. 

\section{Conclusion}
\label{Sec:conclusion}

In this work, we introduced RoboVAD, a benchmark for frame-level anomaly detection in real-world robotic manipulation. RoboVAD enables evaluation across unseen manipulation tasks, unseen anomaly categories and their combination, supporting comparisons between methods using different levels of supervision. We evaluated several existing methods alongside our proposed FlowJigsaw-SSL and FlowJigsaw-Supervised models, and showed that optical flow consistently improves performance over RGB inputs for both variants. Although several methods perform well in-domain, all remain below $70\%$ micro averaged frame-level AUC when both tasks and anomaly categories are unseen. These results highlight the challenges posed by RoboVAD and provide baselines for future research. The current dataset is limited to one robotic platform and a tabletop environment. Future work could expand task and environment diversity and investigate online anomaly detection to support robot recovery.

\bibliographystyle{IEEEtran}
\bibliography{biblio}

\end{document}